\documentclass[]{fairmeta}

\usepackage{amsmath,amsfonts,bm}

\def\eqref#1{equation~\ref{#1}}

\def\1{\bm{1}}

\DeclareMathAlphabet{\mathsfit}{\encodingdefault}{\sfdefault}{m}{sl}
\SetMathAlphabet{\mathsfit}{bold}{\encodingdefault}{\sfdefault}{bx}{n}

\usepackage{amsmath}
\usepackage{amssymb}
\usepackage{array}
\usepackage{booktabs}
\usepackage{colortbl}
\usepackage{enumitem}
\usepackage{fancyhdr}
\usepackage{float}
\usepackage{xurl}

\definecolor{BrandPurple}{HTML}{7366CC}
\definecolor{BrandCyan}{HTML}{00B4E5}
\definecolor{LightGray}{HTML}{F5F5F5}
\definecolor{TableRowAlt}{HTML}{EAF4FB}
\definecolor{HeaderGray}{HTML}{888888}

\newcolumntype{L}[1]{>{\raggedright\arraybackslash}m{#1}}
\newcolumntype{C}[1]{>{\centering\arraybackslash}m{#1}}

\newtcolorbox{highlight}{
    colback=BrandCyan!5,
    colframe=BrandCyan,
    arc=2pt,
    boxrule=0.8pt,
    left=6pt, right=6pt, top=4pt, bottom=4pt,
    breakable
}

\newtcolorbox{keyfinding}{
    colback=BrandPurple!5,
    colframe=BrandPurple,
    coltitle=white,
    fonttitle=\bfseries,
    title=Key Finding,
    arc=2pt,
    boxrule=1pt,
    left=6pt, right=6pt, top=4pt, bottom=4pt,
    breakable
}

\newtcolorbox{tipbox}[1]{
    colback=amapbg,
    colframe=amapblue,
    coltitle=white,
    fonttitle=\bfseries,
    title=#1,
    arc=2pt,
    boxrule=1pt,
    left=6pt, right=6pt, top=4pt, bottom=4pt,
    breakable
}

\newcommand{\model}{DreamX-Phi}
\newcommand{\modelversion}{\model{} 1.0}
\renewcommand{\bibfont}{\footnotesize}
\renewcommand{\headrulewidth}{0pt}
\renewcommand{\footrulewidth}{0pt}

\fancypagestyle{plain}{%
    \fancyhf{}
    \fancyhead[L]{\footnotesize\color{HeaderGray}\sffamily \modelversion{}}
    \fancyhead[R]{\footnotesize\color{HeaderGray} \thepage}
    \fancyfoot[C]{}
    \renewcommand{\headrulewidth}{0pt}
    \renewcommand{\footrulewidth}{0pt}
}

\fancypagestyle{firstpage}{%
    \fancyhf{}
    \fancyhead[L]{\small\sffamily\color{amapblue} DreamX Team}
    \fancyhead[R]{\small\color{HeaderGray} August 2026}
    \fancyfoot[C]{\footnotesize\color{gray}\thepage}
    \renewcommand{\headrulewidth}{0.4pt}
    \renewcommand{\footrulewidth}{0pt}
    \renewcommand{\headrule}{\vspace{-4pt}\color{HeaderGray}\hrule width\headwidth height 0.4pt}
}

\hypersetup{
  pdftitle={\modelversion{}: Action-Conditioned Video World Model for Robotic Manipulation},
  pdfauthor={DreamX Team}
}

\title{\modelversion{}: Action-Conditioned Video World Model for Robotic Manipulation}

\author{DreamX Team}

\abstract{We present \textbf{\modelversion{}}, an action-conditioned video world model for
robotic manipulation that, given an observed frame, a language instruction,
and a prescribed action sequence comprising end-effector poses and gripper
states, predicts the resulting future observations. Yet realism alone does
not guarantee faithfulness: a convincing rollout can still move the wrong arm
or lose the manipulated
object. To ensure the prediction respects each arm's commanded path, we
inject per-arm $\mathrm{SE}(3)$ transformations into attention via
\textbf{PRoPE-style geometric encoding}, preserving arm identity and rigid-motion
structure. Action control alone does not fully constrain scene geometry or the
evolution of small manipulated objects. We therefore add a lightweight \textbf{depth
branch} for scene-level geometry and use \textbf{SAM3 masks} with a frozen \textbf{V-JEPA teacher}
to maintain object consistency throughout grasping. We further distill the
multi-step generator into a few-step student via distribution-matching
distillation for efficient deployment. At the time of writing,
\model{} achieves first place on Track~1 and second place on Track~2 of the
WorldArena~2.0 Challenge. Our model and code will be publicly
available.\footnotemark[1]
}

\metadata[GitHub]{\href{https://github.com/AMAP-ML/DreamX-Phi}{\texttt{github.com/AMAP-ML/DreamX-Phi}}}
\date{August 13, 2026}

\begin{document}

\maketitle
\thispagestyle{firstpage}
\footnotetext[1]{Model weights and inference code will be made publicly available after the
WorldArena~2.0 IROS Challenge concludes.}

\section{Introduction}
\label{sec:introduction}

\begin{figure}[t]
    \centering
    \includegraphics[width=\linewidth]{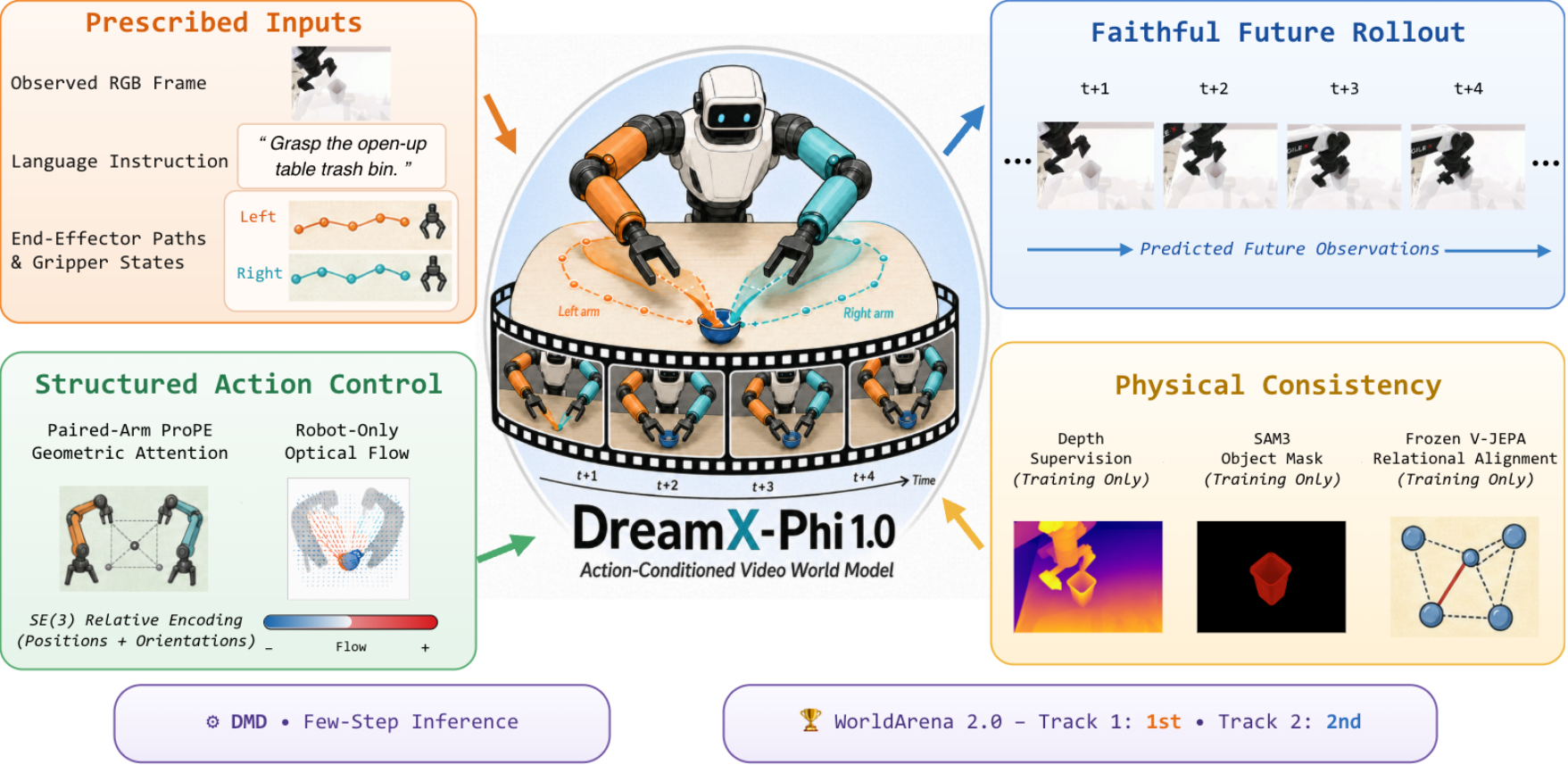}
    \caption{Overview of DreamX-Phi 1.0, an action-conditioned video world model that predicts future observations from a single frame and prescribed bimanual actions.}
    \label{fig:dreamx_phi_overview}
\end{figure}

World models provide a scalable way to evaluate candidate actions before
physical execution. By predicting future scene evolution from the current
observation and a proposed action sequence, they support planning over imagined
outcomes \citep{yang2024unisim,wu2024ivideogpt}. Modern video generators offer
powerful priors over appearance and motion, but photorealistic predictions are
not necessarily faithful to the conditioning actions. Given a fixed initial
observation, distinct prescribed trajectories should induce the corresponding
robot motions and object interactions while leaving action-irrelevant scene
content unchanged. Without this sensitivity, a model may produce a convincing
video yet deviate from the commanded motion, miss the target object, or confuse
grasping with release.

A key challenge is to design an action-conditioned interface that preserves
the spatial structure of robot motion and relates it to the corresponding
changes in the image. Existing approaches commonly encode actions as compact
tokens or feature-wise modulation
\citep{wu2024ivideogpt,zhu2025irasim,huang2025vid2world}. While compact, these
representations do not explicitly preserve the rigid-body geometry of
end-effector motion or indicate where the commanded motion should appear in
the generated video. A more structured representation can encode both aspects
explicitly: $\mathrm{SE}(3)$ trajectories describe how the robot moves in 3D,
while dense motion cues indicate where and how that motion appears in the image
\citep{miyato2024gta,li2025cameras,chen2026flowwam}. Correct robot motion alone,
however, does not guarantee a faithful rollout; prediction fidelity also
depends on preserving scene geometry and the state of the manipulated object
as the interaction unfolds. This motivates combining geometry-aware action
conditioning with dense supervision of the resulting scene evolution.

To address these challenges, we introduce \modelversion{}, a geometry-aware,
action-conditioned video world model based on Wan2.2-TI2V-5B
\citep{wan2025,wan2025wan22ti2v5b}. Rather than compressing the prescribed
action sequence into a generic control embedding, DreamX-Phi preserves the
$\mathrm{SE}(3)$ structure of end-effector motion through PRoPE-based attention
and relates it to the corresponding image dynamics through action-derived
motion cues \citep{miyato2024gta,li2025cameras,chen2026flowwam}. Its learning
objective captures the physical consequences of interaction by regularizing
scene geometry and object-centric temporal consistency. This encourages
action-faithful rollouts in which the manipulated object retains its visual
identity and evolves coherently over time. For efficient deployment, DMD
post-training distills the iterative generation process into a few-step sampler
\citep{yin2024onestep,yin2024improved}.

To support both general video prediction and action-conditioned dynamics, we
curate a heterogeneous training corpus from egocentric video and robot
interaction data spanning real and simulated environments, as detailed in
Section~\ref{sec:training}. Action-agnostic clips broaden the model's visual and
motion priors, while temporally aligned robot trajectories ground future
predictions in control. For RoboTwin, we process the training videos with DreamX-Refiner to improve
their resolution and visual quality. Evaluations on
WorldArena~1.0 and 2.0 demonstrate strong performance
\citep{shang2026worldarena,shang2026worldarena2}. In the fixed WorldArena~2.0
leaderboard snapshot, our submissions rank first on Track~1 and tie for second
on Track~2. DreamX-Phi also achieves an offline EWMScore-P of 76.88 on
WorldArena~1.0 Track~1.

Our contributions are threefold:
\begin{itemize}[leftmargin=*]
    \item We introduce a geometry-aware action representation that connects
    prescribed robot actions to their visual consequences. By combining
    structured $\mathrm{SE}(3)$ trajectories with image-space motion cues, it
    encodes both the commanded 3D end-effector motion and where its effects
    should appear in the predicted video.
    \item We propose manipulation-aware supervision that focuses learning on
    scene geometry and the state of the manipulated object. By emphasizing
    depth structure and object evolution, it encourages physically coherent
    robot--object interactions throughout the predicted rollout.
    \item Extensive evaluations on WorldArena~1.0 and 2.0 demonstrate the
    effectiveness of DreamX-Phi. In the fixed WorldArena~2.0 snapshot, our
    submissions rank first on Track~1 and tie for second on
    Track~2; our model also achieves an offline EWMScore-P of 76.88 on
    WorldArena~1.0 Track~1.
\end{itemize}

\section{Related Work}
\label{sec:related_work}

\paragraph{\textbf{Action-Conditioned Video World Models.}}
Large video generators such as Wan and Cosmos~3 provide scalable visual priors
for physical dynamics \citep{wan2025,nvidia2026cosmos3}. Building on such
priors, interactive models predict how a scene evolves under an external
control signal. UniSim, iVideoGPT, and DreamX-World study controllable visual
dynamics across broad domains, while AVID adapts pretrained video diffusion
through a learned action interface
\citep{yang2024unisim,wu2024ivideogpt,dreamxteam2026world,rigter2025avid}.
Robot-specific models make the conditioning contract more concrete: IRASim
aligns robot trajectories with video frames, Vid2World introduces causal
generation with action guidance, and HMA learns heterogeneous action--video dynamics
\citep{zhu2025irasim,huang2025vid2world,wang2025hma}. These methods establish
video prediction as a learned simulator, but action fidelity remains difficult:
a visually plausible rollout may still move the wrong arm or produce an
incorrect object response. \model{} focuses on this prescribed-action setting,
predicting future observations from a given bimanual trajectory rather than
generating the actions themselves.

\paragraph{World Action Models for Robotics.}
Current world action models (WAMs) connect video and control in three main
ways. The first injects low-dimensional action tokens or adapters into a video
generator, as in IRASim, Vid2World, and HMA
\citep{zhu2025irasim,huang2025vid2world,wang2025hma}. The second jointly models
visual futures and actions, enabling the video model to act as a policy or
planner, as in UVA, WorldVLA, LingBot-VA, DreamZero, and Cosmos Policy
\citep{li2025uva,cen2025worldvla,li2026lingbotva,ye2026dreamzero,
kim2026cosmospolicy}. The third converts robot motion into a spatially aligned
condition: OSCAR renders kinematic skeletons, Robot-Factored World Models
render robot geometry from commands, and FlowWAM represents actions with
optical flow \citep{wu2026oscar,kim2026robotfactored,chen2026flowwam}. These
interfaces trade compactness for structure: token-based controls are general
but geometrically implicit, whereas rendered or flow-based controls localize
motion in the image but do not directly preserve the continuous rigid-body
trajectory of each arm.

\paragraph{Structured Control and Physical-Consistency Supervision.}
Geometry-aware attention provides a direct way to retain rigid-motion
structure. GTA inserts relative $\mathrm{SE}(3)$ transformations into
attention, while projective relative positional encoding applies known camera
geometry to queries, keys, values, and outputs
\citep{miyato2024gta,li2025cameras}. Geometry alone, however, constrains the
commanded robot motion rather than the full scene response. Depth supervision
can organize scene geometry, object masks can prevent small contact regions
from being overwhelmed by a uniform generative objective
\citep{lipman2023flow}, and predictive video features can regularize object evolution
\citep{guo2026xwam,carion2025sam3segmentconcepts,assran2025vjepa2}. \model{}
combines these complementary signals: arm-specific PRoPE preserves the
continuous $\mathrm{SE}(3)$ trajectory, robot-only flow supplies an
image-aligned motion cue, and depth, SAM3 mask weighting, and frozen V-JEPA
relational supervision target scene and object consistency. The central design
choice is therefore to preserve the structure of the commanded action while
separately supervising its visual and physical consequences.

\section{Data Curation}
\label{sec:training}

\paragraph{Data Sources.}
Reliable action-conditioned prediction depends not only on the diversity of
robot motions, but also on broad visual coverage and consistent alignment
between observations and control signals. We therefore construct the corpus
from three complementary sources: action-free egocentric video, real-robot
demonstrations, and simulated robot trajectories. Together, these sources span
everyday visual dynamics, physically executed manipulation, and controlled
variation in task and scene configuration. Table~\ref{tab:data_sources}
summarizes their scale. Throughout curation, we preserve the camera structure
provided by each source: single-view recordings remain single-view examples,
whereas synchronized multi-view observations are combined as described below.

\paragraph{Curation and Normalization.}
The raw robot collections contain behaviors that are poorly matched to the
manipulation setting studied in this report. We remove trajectories dominated
by mobile-base motion, dexterous-hand operation, or stationary segments, while
deliberately retaining failed task executions because they expose informative
failure modes and non-ideal interaction dynamics. After removing mobile-base
and stationary segments, the filtered AgiBot imitation-learning split contains
178.7 hours. For sources with action annotations, we normalize observations,
instructions, robot states, and actions into a common LeRobot v2.1
representation, providing a consistent interface across otherwise
heterogeneous datasets.

\begin{table}[H]
    \centering
    \small
    \setlength{\tabcolsep}{3pt}
    \caption{Data sources used to construct the curated corpus. RoboTwin is
    reported in clips because its duration is not available.}
    \label{tab:data_sources}
    \begin{tabular}{@{}L{0.34\linewidth} L{0.25\linewidth} L{0.29\linewidth}@{}}
        \toprule
        Source & Domain & Data volume \\
        \midrule
        Ego4D \citep{grauman2022ego4d}
            & Egocentric video & 3,700 h \\
        AgiBot World 2026 \citep{agibotworld2026}
            & Real robot & 1,900 h \\
        InternData-A1 \citep{tian2025interndata}
            & Real / simulated robot & 78 h real; 3,747 h simulated \\
        Cosmos3-DROID \citep{khazatsky2024droid,nvidia2026cosmos3droid}
            & Real robot & 350 h \\
        RoboCOIN \citep{wu2025robocoin}
            & Real robot & 618 h \\
        RoboTwin 2.0 \citep{chen2025robotwin2}
            & Simulated robot & 25,000 action-annotated clips \\
        \bottomrule
    \end{tabular}
\end{table}

\paragraph{Action-Agnostic Pretraining.}
The action-agnostic pool includes every retained video, regardless of whether
the source also provides action annotations. A single camera stream remains a
single-view example; when synchronized streams from multiple cameras are
available, we spatially concatenate them into a unified multi-view video. This
view-adaptive organization preserves the information available in each source
and makes the curated corpus applicable to both single-view and multi-view
downstream settings. It also exposes the model to the complementary appearance
statistics and motion patterns found in egocentric video, real-robot operation,
and simulation.

\paragraph{Action-Conditioned Fine-Tuning.}
The action-conditioned pool is restricted to videos with synchronized action
annotations. Each video is paired with its corresponding robot trajectory and
the annotations available in the common representation, so that visual and
control streams remain temporally consistent. Its RoboTwin component contains
25,000 bimanual clips, spanning both clean and randomized variants. Before
these clips enter the curated pool, we apply our video refinement model,
DreamX-Refiner, to super-resolve the RoboTwin videos; the resulting
high-resolution clips provide the visual data used in this phase.

\section{Method}
\label{sec:method}

\subsection{Overview}

\begin{figure}[t]
    \centering
    \includegraphics[width=\linewidth]{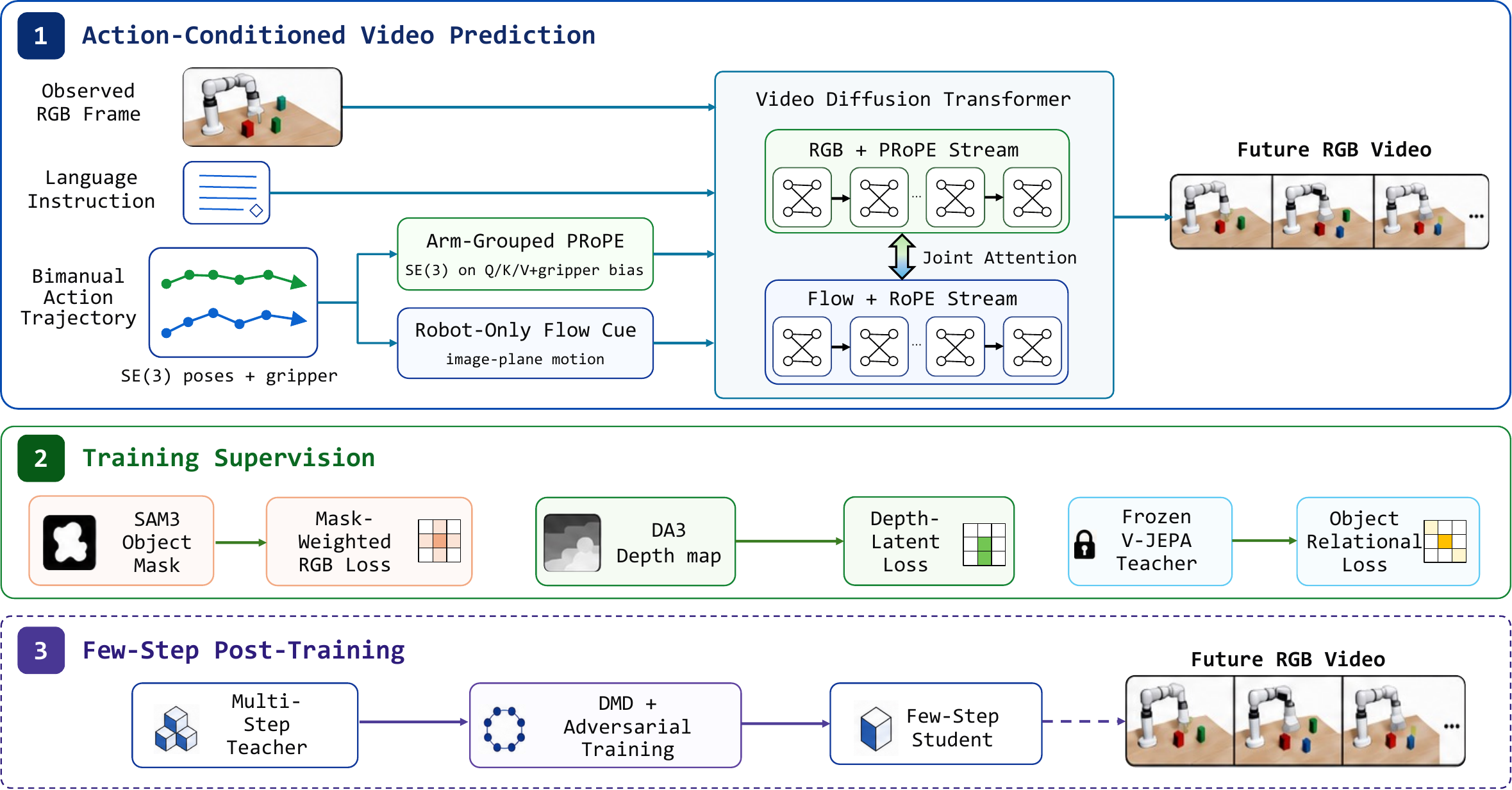}
    \caption{Overview of the \modelversion{} framework. (1) Given an observed
    RGB frame, a language instruction, and an externally specified bimanual
    action trajectory, a Wan2.2-TI2V-5B video diffusion transformer predicts
    the future RGB video. Arm-grouped PRoPE and a robot-only optical-flow cue
    provide complementary geometric and image-plane action conditioning. (2)
    During training, SAM3-derived masks reweight the RGB objective, Depth Anything~3 (DA3) depth maps provide targets for the depth-latent objective, and a frozen
    V-JEPA teacher supplies object-relational supervision. (3) DMD
    with adversarial training distills the multi-step teacher into a few-step
    student.}
    \label{fig:dreamx_phi_pipeline}
\end{figure}

Given an observed RGB frame $\mathbf{x}_{0}$, a language instruction
$\mathbf{c}$, and a prescribed bimanual action trajectory
$\mathbf{a}_{1:T}$ containing end-effector poses and gripper states, our goal
is to model the conditional distribution
\begin{equation}
    p_{\theta}(\mathbf{x}_{1:T}\mid
    \mathbf{x}_{0},\mathbf{a}_{1:T},\mathbf{c})
\end{equation}
using a Wan2.2-TI2V-5B video diffusion transformer. The latent of the first
frame provides the visual context, and the future-frame latents are learned
under a flow-matching objective \citep{lipman2023flow}. The resulting design
combines structured action conditioning with auxiliary geometric and
object-centric supervision, while retaining DMD as a post-training
route to few-step inference. As illustrated in Figure~\ref{fig:dreamx_phi_pipeline}, the framework is
organized into three parts: action-conditioned video prediction, training
supervision, and few-step post-training.

\subsection{PRoPE Control for Bimanual Actions}
\label{sec:prope}

World action models commonly encode actions as low-dimensional tokens or
additive features and inject them through concatenation, modulation, or
cross-attention
\citep{wu2024ivideogpt,zhu2025irasim,huang2025vid2world,nvidia2026cosmos3}.
These generic interfaces are flexible, but they leave the rigid-body relations
within an end-effector trajectory to be inferred implicitly. PRoPE provides a
more structured alternative by inserting known relative transforms directly
into self-attention. The transforms affect both attention weights and value
aggregation while remaining invariant to the global coordinate frame
\citep{li2025cameras}. We incorporate this mechanism as a residual branch, so
geometric control augments rather than replaces the pretrained generative path
\citep{dreamxteam2026world}. Because robot end-effector commands form ordered
$\mathrm{SE}(3)$ trajectories, this group-action formulation is a natural
interface for action conditioning.

Our use of PRoPE differs from its original camera setting: an end effector is
not treated as a physical camera. Instead, we reuse only the group-action
attention mechanism to expose relative rigid transforms along an ordered robot
trajectory. Adapting the camera formulation requires three corresponding
changes: all arms must be expressed in a shared coordinate system, each arm
must retain a persistent portion of the attention representation, and the
gripper state must be injected separately from the $\mathrm{SE}(3)$
transform.

\paragraph{Action Representation.}
At frame $t$, arm $k$ is described by position $\mathbf{p}_t^k$, quaternion
$\mathbf{q}_t^k$ (converted to $\mathbf{R}_t^k$), and gripper value
$g_t^k$. We first construct its end-effector frame and then express that frame
relative to the initial pose of arm~1:
\begin{equation}
    \mathbf{G}_t^k=
    \begin{bmatrix}
        \mathbf{R}_t^k & \mathbf{p}_t^k \\
        \mathbf{0}^{\top} & 1
    \end{bmatrix},
    \qquad
    \bar{\mathbf{G}}_t^k=(\mathbf{G}_1^1)^{-1}\mathbf{G}_t^k.
    \label{eq:ee_pose}
\end{equation}
This construction places all arms in a common reference frame. We then
normalize translations using a single motion-amplitude factor,
\begin{equation}
    \gamma=\max_{k,t}
    \left\lVert\bar{\mathbf{p}}_t^k-\bar{\mathbf{p}}_1^k\right\rVert_2,
    \qquad
    s_{\gamma}=\begin{cases}
        \gamma, & \gamma>\epsilon,\\
        1, & \gamma\leq\epsilon,
    \end{cases}
    \qquad
    \widetilde{\mathbf{G}}_t^k=
    \begin{bmatrix}
        \bar{\mathbf{R}}_t^k & \bar{\mathbf{p}}_t^k/s_{\gamma} \\
        \mathbf{0}^{\top} & 1
    \end{bmatrix}.
    \label{eq:scaled_pose}
\end{equation}
Because $\gamma$ measures motion amplitude rather than absolute workspace
size, the resting distance between the arms does not dominate the scale. We
next invert the normalized frames,
$\mathbf{A}_t^k=(\widetilde{\mathbf{G}}_t^k)^{-1}$, and temporally align both
$\mathbf{A}_t^k$ and $g_t^k$ with the VAE latent frames. In the two-arm
setting, this yields
$\mathbf{A}\in\mathbb{R}^{2\times T_{\mathrm{lat}}\times4\times4}$ and
$\mathbf{g}\in\mathbb{R}^{2\times T_{\mathrm{lat}}}$. A missing arm is
represented by identity poses with $g_t^k=0$.

\paragraph{Geometric Attention.}
Each transformer block contains a parallel attention branch with dedicated
query, key, value, and output projections, conditioned on $\mathbf{A}$ and
$\mathbf{g}$. We adopt an identity intrinsic matrix,
$\mathbf{K}=\mathbf{I}_3$, so the PRoPE projection matrix reduces to
$\mathbf{P}_t^k=\mathbf{A}_t^k$
\citep{li2025cameras}. We partition the attention heads into fixed contiguous
groups $\{\mathcal{H}_k\}$, with one group assigned to each arm. For token $i$
at latent frame $n(i)$ and head $h\in\mathcal{H}_k$, we define
$\mathbf{D}_i=\mathbf{I}_{d_h/4}\otimes\mathbf{A}_{n(i)}^k$ and apply the
following token-wise transforms:
\begin{equation}
    \mathbf{Q}'_i=\mathbf{D}_i^{\top}\mathbf{Q}_i,
    \qquad
    \mathbf{K}'_i=\mathbf{D}_i^{-1}\mathbf{K}_i,
    \qquad
    \mathbf{V}'_i=\mathbf{D}_i^{-1}\mathbf{V}_i,
    \qquad
    \mathbf{O}^{\mathrm{act}}_i=
    \mathbf{D}_i\left[
        \operatorname{Attn}(\mathbf{Q}',\mathbf{K}',\mathbf{V}')
    \right]_i.
    \label{eq:prope_attn}
\end{equation}
All patches associated with the same frame and arm share $\mathbf{D}_i$.
Consequently, a token pair $(i,j)$ is coupled through the relative motion
$\mathbf{D}_i\mathbf{D}_j^{-1}$ rather than through an absolute coordinate
frame.

\paragraph{Gripper and Residual.}
Gripper opening is scalar-valued and therefore cannot be represented as an
$\mathrm{SE}(3)$ element. We inject it after the inverse geometric map as a
per-arm bias on the corresponding attention heads:
\begin{equation}
    \mathbf{b}_t^k=\mathbf{W}_g g_t^k+\mathbf{b}_g,
    \qquad
    \mathbf{o}_{t,u,h}^{\mathrm{act}}
    \leftarrow\mathbf{o}_{t,u,h}^{\mathrm{act}}+\mathbf{b}_t^k,
    \quad h\in\mathcal{H}_k,
    \label{eq:gripper_bias}
\end{equation}
broadcast over the spatial locations $u$ of all heads in $\mathcal{H}_k$.
The resulting heads are concatenated, projected back to the model width, and
added to the pretrained self-attention output. Both the gripper adapter and
this output projection are initialized to zero, keeping the residual branch
silent until it is updated during training.


\subsection{Auxiliary Depth Supervision for 3D Consistency}

An RGB prediction objective can capture appearance and motion without
explicitly constraining surface ordering, object extent, or contact geometry.
To supply this missing geometric signal during training, we draw on the depth
adaptation design of X-WAM \citep{guo2026xwam} and add a lightweight auxiliary
depth branch. Let $\mathbf{d}$ denote a depth video aligned with the RGB
sequence. We replicate each single-channel depth map across the channel
dimension to obtain a pseudo-RGB input, then encode it with the same frozen
video VAE used for RGB. This produces the latent depth target
$\mathbf{z}^{d}=\mathcal{E}(\mathbf{d})$, which is predicted by a branch
attached to the tail of the RGB transformer.

Concretely, for an RGB transformer with $N$ blocks, we replicate its final
$M$ blocks ($M<N$) to form the auxiliary branch, leaving the first $N-M$
blocks as a shared trunk. The trunk output initializes both pathways, and each
replicated depth block is initialized from its pretrained RGB counterpart. At
every adapted layer $j$, cross-attention allows the depth pathway to read the
corresponding RGB representation:
\begin{equation}
    \mathbf{h}^{j}_{d}=
    \operatorname{DepthBlock}_{j}\!\left(
    \mathbf{h}^{j-1}_{d};
    \mathbf{K}^{j}_{\mathrm{rgb}},
    \mathbf{V}^{j}_{\mathrm{rgb}}\right),
    \qquad j=1,\ldots,M .
\end{equation}
Here, $\mathbf{K}^{j}_{\mathrm{rgb}}$ and
$\mathbf{V}^{j}_{\mathrm{rgb}}$ are obtained from the RGB branch at layer
$j$. The connection is deliberately one-way: the depth branch can consume RGB
features, but the RGB branch never consumes depth features. This asymmetric
design leaves the RGB forward computation unchanged and therefore keeps depth
prediction optional at inference.

A dedicated output head maps the final depth tokens to
$\widehat{\mathbf{z}}^{d}$. We supervise this prediction with a latent-space
mean-squared error,
\begin{equation}
    \mathcal{L}_{\mathrm{depth}} =
    \frac{1}{|\mathbf{z}^{d}|}
    \left\|\widehat{\mathbf{z}}^{d}-\mathbf{z}^{d}\right\|_2^2 .
\end{equation}
Thus, unlike the RGB generation pathway, the depth branch is supervised
directly in latent space rather than treated as a separate noisy diffusion
sequence. The auxiliary objective encourages the shared representation to
encode stronger geometric structure during training, without introducing a
depth-related requirement at deployment.

\subsection{Object-Centric Physical Consistency}
\label{sec:object}

The flow-matching RGB objective averages errors over all valid future tokens.
In manipulation videos, however, the robot arm and manipulated object often
occupy only a small fraction of the frame, allowing the static background to
dominate the contact-local errors that determine whether an interaction is
physically plausible. A rollout can therefore appear photorealistic even when
the gripper misses or penetrates the object, the object does not respond to
contact, or its shape and state change abruptly after a grasp. PRoPE constrains
the commanded arm motion but, on its own, does not enforce a coherent object
response. To address these complementary aspects, we use a SAM3 mask to focus
the RGB objective on the manipulated object and a frozen V-JEPA teacher to
regularize its spatiotemporal evolution
\citep{carion2025sam3segmentconcepts,assran2025vjepa2}. Together, these signals
encourage predicted object motion to remain coupled to arm contact, rather
than rewarding global visual quality alone.

\paragraph{Object-Aware Supervision.}
Offline SAM3 processing provides a binary mask video for the manipulated
object. The mask is used only for supervision: SAM3 is not fine-tuned jointly
with the model, and no mask is required at inference. After projection onto
the latent grid, token $i$ is assigned $m_i\in\{0,1\}$, and its normalized
weight is defined as
\begin{equation}
    \widetilde{w}_i = 1+(\lambda_m-1)m_i,
    \qquad
    w_i = \frac{\widetilde{w}_i}
    {\frac{1}{|\mathcal{V}|}\sum_{j\in\mathcal{V}}\widetilde{w}_j},
    \qquad
    \mathcal{L}_{\mathrm{rgb}}^{\mathrm{obj}} =
    \frac{1}{|\mathcal{V}|}
    \sum_{i\in\mathcal{V}}w_i\ell_i^{\mathrm{FM}},
    \label{eq:obj_rgb}
\end{equation}
where $\mathcal{V}$ denotes the set of valid future tokens and
$\lambda_m>1$ is the object-to-background ratio before normalization. The
mean-weight normalization stabilizes the overall loss scale as the mask area
changes, while clips without a valid mask retain uniform weights. As a result,
contact-local errors---such as a missed object displacement or a
contact-induced deformation---remain influential despite the much larger
static background.

\paragraph{V-JEPA Alignment for Physical Consistency.}
Mask reweighting operates on local flow-matching errors and does not, by
itself, determine whether an object follows a coherent trajectory through
contact. Similar frame-level errors may still conceal temporal drift, an
inconsistent grasp state, or object motion that is decoupled from the arm. We
therefore use a frozen V-JEPA teacher to constrain relations among object
features across both space and time. For each sample $b$, we select a
temporally stratified set of masked teacher tokens $\mathcal{I}_b$, capped at
$M_{\max}$, and interpolate the corresponding video-model hidden tokens to the
same coordinates. Let $M_b=|\mathcal{I}_b|$, and let the normalized projected
student and teacher features be
$\mathbf{S}_b,\mathbf{Q}_b\in\mathbb{R}^{M_b\times d}$. Rather than matching
feature coordinates directly, we align their Gram matrices so that the
student is not tied to the teacher's feature basis:
\begin{equation}
    \ell_{\mathrm{JEPA}}^{(b)} =
    \frac{1}{M_b^2}\left\|
    \mathbf{S}_b\mathbf{S}_b^{\top}
    -\mathbf{Q}_b\mathbf{Q}_b^{\top}
    \right\|_1.
    \label{eq:jepa}
\end{equation}
The pairwise relational objective discourages object identity, shape, and
state from drifting across the contact interval. In this sense, the
mask-weighted RGB term identifies where prediction accuracy is most important,
whereas V-JEPA constrains how the manipulated object evolves over time; the
two objectives jointly promote physically consistent arm--object interaction.
To keep this supervision stable, a sample contributes to the relational loss
only when its mask provides enough tokens and its flow-matching noise is not
too large. Specifically,
\begin{equation}
    r_b=
    \mathbb{I}[M_b\geq M_{\min}]
    \mathbb{I}[\sigma_b\leq\sigma_{\max}],
    \qquad
    \mathcal{L}_{\mathrm{JEPA}}=
    \frac{\sum_b r_b\ell_{\mathrm{JEPA}}^{(b)}}
    {\max(1,\sum_b r_b)},
    \label{eq:jepa_gate}
\end{equation}
Samples that fail either gate contribute zero. The teacher remains frozen
throughout training. For eligible samples, the projector receives gradients;
during an initial projector-only phase, gradients stop at the video-model
hidden state and are subsequently opened linearly to the trainable video-model
parameters.

\subsection{Few-Step Post-Training}

To reduce the number of denoising evaluations, we distill the multi-step generator following DMD2 \citep{yin2024improved}. Let \(G_{\eta}\) denote the \(N\)-step student,
let \(\mathbf{y}=(\mathbf{x}_{0},\mathbf{a}_{1:T},\mathbf{c})\) collect its
conditions, and let \(\widetilde{\mathbf{z}}_{0}\) be the student's clean
future-video latent prediction at a sampled denoising step. In contrast to the
text-conditioned image-generation setting, \(\mathbf{y}\) includes both the
observed frame and the temporally aligned, prescribed bimanual action
trajectory. The student, frozen teacher, and online fake-score denoiser all
receive the same \(\mathbf{y}\); the adversarial classification head instead
operates on the denoiser's bottleneck features. 

For a noise level \(\tau\) sampled from the DMD distribution, let
\(q_{\eta,\tau}(\cdot\mid\mathbf{y})\) and
\(p_{\mathrm{data},\tau}(\cdot\mid\mathbf{y})\) denote the conditional
marginals produced by applying the same Wan forward-noising process to the
student and real future-video latents, respectively. Distribution matching is
then expressed as the following KL objective:
\begin{equation}
\mathcal{L}_{\mathrm{DMD}}(\eta)
=\mathbb{E}_{\mathbf{y}\sim p_{\mathrm{data}}(\mathbf{y}),\,\tau}\!\left[
D_{\mathrm{KL}}\!\left(
q_{\eta,\tau}(\cdot\mid\mathbf{y})
\,\|\,
p_{\mathrm{data},\tau}(\cdot\mid\mathbf{y})
\right)\right].
\label{eq:dmd-objective}
\end{equation}

We complement this distribution-matching term with the noised
non-saturating GAN objective. Given
\((\mathbf{z}_{0}^{\star},\mathbf{y})\sim p_{\mathrm{data}}\) and a sampled GAN
noise level \(u\), we obtain \(\mathbf{z}_{u}^{r}\) and
\(\mathbf{z}_{u}^{f}\) by independently applying the same Wan forward-noising
process at \(u\) to \(\mathbf{z}_{0}^{\star}\) and
\(\widetilde{\mathbf{z}}_{0}\), respectively. Let
\(D(\cdot,u;\mathbf{y})\in(0,1)\) denote the classification head's conditional
probability that a latent is real. The generator and discriminator objectives
are
\begin{equation}
\begin{aligned}
\mathcal{L}_{\mathrm{adv}}^{G}
&=-\mathbb{E}\!\left[
\log D(\mathbf{z}_{u}^{f},u;\mathbf{y})\right],\\
\mathcal{L}_{\mathrm{adv}}^{D}
&=-\mathbb{E}\!\left[
\log D(\mathbf{z}_{u}^{r},u;\mathbf{y})
+\log\!\left(1-D
(\mathbf{z}_{u}^{f},u;\mathbf{y})\right)\right].
\end{aligned}
\label{eq:dmd-adversarial-objectives}
\end{equation}
The student is therefore optimized with
\begin{equation}
\mathcal{L}_{\mathrm{student}}^{\mathrm{DMD}}
=\mathcal{L}_{\mathrm{DMD}}
+\lambda_{\mathrm{adv}}\mathcal{L}_{\mathrm{adv}}^{G},
\qquad \lambda_{\mathrm{adv}}\geq 0.
\label{eq:dmd-total-objective}
\end{equation}
The fake-score denoiser and its adversarial classification head are updated in
a separate auxiliary step. Finally, the few-step student uses the same fixed
\(N\)-step denoising schedule during training and inference. Under backward
simulation, earlier student steps generate the input to a sampled step without
gradient tracking; the conditioning tuple \(\mathbf{y}\) remains fixed across
all steps, and gradients propagate only through the sampled step.

\section{Evaluation}
\label{sec:evaluation}

We evaluate DreamX-Phi 1.0 along two complementary axes: the fidelity of its
predicted visual rollouts and its utility as a learned environment for policy
training. WorldArena~2.0 is our primary benchmark: Track~1 evaluates video
prediction conditioned on language instructions or robot actions, while
Track~2 measures whether a policy optimized through interaction with the
submitted world model succeeds in held-out simulator episodes. We additionally report
WorldArena~1.0 Track~1 results to contextualize performance against earlier
world models \citep{shang2026worldarena,shang2026worldarena2}.

\subsection{Datasets}

Both benchmarks use evaluation sets curated and released by the WorldArena
organizers from RoboTwin~2.0 trajectories
\citep{chen2025robotwin2,shang2026worldarena,shang2026worldarena2}.
WorldArena~2.0 Track~1 contains 1,000 episodes. Each provides an initial RGB
observation together with a language instruction and a robot action trajectory,
and the model predicts the subsequent rollout conditioned on either signal.
WorldArena~1.0 Track~1 follows the Clean-50 protocol, covering 50 manipulation
tasks with 10 held-out episodes per task.

WorldArena~2.0 Track~2 evaluates whether the learned dynamics are useful beyond
open-loop video prediction. The submitted world model serves as the rollout
environment for optimizing a $\pi_{0.5}$ policy using an organizer-provided
initialization and a fixed reward model \citep{physicalintelligence2025pi05}.
The resulting policy is then evaluated on held-out Adjust Bottle episodes in
RoboTwin~2.0 \citep{shang2026worldarena2}.

\subsection{Metrics}

For Track~1, we report EWMScore-P together with the 15 normalized component
metrics exposed by the official leaderboards. The metrics span visual quality,
temporal dynamics, content consistency, physical interaction, 3D structure,
and conditioning fidelity, while EWMScore-P summarizes overall performance by
averaging the component scores. WorldArena~2.0 additionally caps Dynamic
Degree, Flow Score, and Motion Smoothness by their ground-truth reference
values before aggregation. Track~2 is evaluated by the policy success rate on
held-out Adjust Bottle episodes.

\subsection{WorldArena~2.0 Results}

Because the leaderboard is continuously updated, all WorldArena~2.0 results in
this report are taken from the official snapshot at commit
\texttt{cb8f9c2}, dated
August~12, 2026.\footnote{Leaderboard comparisons are anchored to the official
\href{https://huggingface.co/spaces/WorldArena/WorldArena2.0/commit/cb8f9c239a302fa283c472269b7b86d4754f8992}{WorldArena~2.0 snapshot at commit \texttt{cb8f9c2}}
and the official
\href{https://huggingface.co/spaces/WorldArena/WorldArena/commit/483dfcc9a2d57974809c15d2a235c3e204f2b0f2}{WorldArena~1.0 snapshot at commit \texttt{483dfcc}}.}
We report both tracks under the checkpoint name DreamX-Phi-1.0-FDM-0730.\footnote{The
official Track~1 and Track~2 artifacts list the corresponding submission
identifiers as \texttt{JF\_World} and \texttt{DreamX-Phi}, respectively.} We report
the complete leaderboard Top~3 together with selected open-source reference
models evaluated in the benchmark paper; all displayed values come from the
leaderboard snapshot rather than the paper tables.
Tables~\ref{tab:worldarena2-track1}
and~\ref{tab:worldarena2-track2} report the two tracks, respectively.

\begin{table}[H]
    \centering
    \caption{WorldArena~2.0 Track~1 leaderboard comparison at the
    August~12, 2026 snapshot. We report the official Top~3 and selected
    open-source reference systems. Scores use a 0--100 scale, and EWMScore-P
    averages the 15 component scores. Boldface and underlining denote the best
    and second-best results among the displayed systems, respectively.}
    \label{tab:worldarena2-track1}
    \scriptsize
    \setlength{\tabcolsep}{2.1pt}
    \renewcommand{\arraystretch}{1.08}

    \textbf{(a) Visual quality, motion quality, and content consistency}
    \par\vspace{2pt}
    \begin{tabular}{@{}L{2.65cm}*{9}{r}@{}}
        \toprule
        \multirow{2}{*}{Model}
        & \multicolumn{3}{c}{Visual Quality}
        & \multicolumn{3}{c}{Motion Quality}
        & \multicolumn{3}{c}{Content Consistency} \\
        \cmidrule(lr){2-4}\cmidrule(lr){5-7}\cmidrule(lr){8-10}
        & \shortstack{Image\\Quality} & \shortstack{Aesthetic\\Quality}
        & \shortstack{JEPA\\Similarity} & \shortstack{Dynamic\\Degree}
        & \shortstack{Flow\\Score} & \shortstack{Motion\\Smoothness}
        & \shortstack{Subject\\Consistency} & \shortstack{Background\\Consistency}
        & \shortstack{Photometric\\Consistency} \\
        \midrule
        \rowcolor{BrandCyan!8}
        \shortstack[l]{\textbf{DreamX-Phi-}\\
        \textbf{1.0-FDM-0730}}
        & 63.25 & 43.38
        & \underline{92.93} & \underline{22.90} & \textbf{5.81} & 63.26
        & \underline{71.93} & \underline{84.52} & 14.29 \\
        Alpha-World & \textbf{64.59} & \underline{43.83} & \textbf{92.96}
        & 22.83 & \underline{5.80} & 63.13 & 71.76 & 83.85 & \underline{14.84} \\
        FlowWAM-FiveAges
        & \underline{64.31} & \textbf{44.00} & 92.54 & 22.76 & \underline{5.80} & 62.67
        & \textbf{71.94} & 83.69 & \textbf{14.96} \\
        GigaWorld-0 & 50.57 & 43.30 & 44.41 & 22.59 & 5.79 & 62.73
        & 56.11 & 79.77 & 3.60 \\
        Vidar & 45.33 & 42.75 & 54.76 & 18.16 & 5.60 & \underline{63.63}
        & 53.36 & 69.95 & 10.96 \\
        WoW & 49.37 & 40.63 & 73.25 & 22.27 & 5.76 & 62.28
        & 65.55 & 83.41 & 10.73 \\
        Ctrl-World & 53.78 & 38.65 & 78.70 & \textbf{23.15} & 5.79
        & \textbf{63.70} & 70.63 & \textbf{85.05} & 14.21 \\
        IRASim & 45.18 & 31.43 & 62.15 & 18.44 & 5.49 & 59.64
        & 48.02 & 65.70 & 10.44 \\
        \bottomrule
    \end{tabular}

    \vspace{0.8em}
    \textbf{(b) Physics adherence, 3D accuracy, controllability, and overall score}
    \par\vspace{2pt}
    \begin{tabular}{@{}L{2.65cm}*{7}{r}@{}}
        \toprule
        \multirow{2}{*}{Model}
        & \multicolumn{2}{c}{Physics Adherence}
        & \multicolumn{2}{c}{3D Accuracy}
        & \multicolumn{2}{c}{Controllability}
        & \multirow{2}{*}{\shortstack{EWM\\Score-P}} \\
        \cmidrule(lr){2-3}\cmidrule(lr){4-5}\cmidrule(lr){6-7}
        & \shortstack{Interaction\\Quality} & \shortstack{Trajectory\\Accuracy}
        & \shortstack{Depth\\Accuracy} & Perspectivity
        & \shortstack{Instruction\\Following} & \shortstack{Semantic\\Alignment}
        & \\
        \midrule
        \rowcolor{BrandCyan!8}
        \shortstack[l]{\textbf{DreamX-Phi-}\\
        \textbf{1.0-FDM-0730}}
        & \textbf{57.36} & \textbf{57.15}
        & \underline{98.55} & \underline{82.24} & \textbf{61.62}
        & 90.53 & \textbf{60.65} \\
        Alpha-World & \underline{57.18} & 49.22 & 97.14 & \textbf{82.38}
        & \underline{60.58} & \underline{91.84} & \underline{60.13} \\
        FlowWAM-FiveAges
        & 53.74 & \underline{49.76} & \textbf{98.99} & 80.52 & 58.06
        & \textbf{92.06} & 59.72 \\
        GigaWorld-0 & 47.46 & 15.24 & 77.22 & 74.98 & 51.22 & 85.92 & 48.06 \\
        Vidar & 43.04 & 16.99 & 81.55 & 67.70 & 45.96 & 87.24 & 47.13 \\
        WoW & 49.87 & 21.39 & 80.08 & 76.24 & 52.95 & 87.67 & 52.10 \\
        Ctrl-World & 52.72 & 45.24 & 94.50 & 76.56 & 52.52 & 88.45 & 56.24 \\
        IRASim & 37.88 & 22.68 & 89.68 & 51.56 & 38.18 & 88.03 & 44.97 \\
        \bottomrule
    \end{tabular}
\end{table}

\begin{table}[H]
    \centering
    \caption{WorldArena~2.0 Track~2 results at the August~12, 2026 snapshot.
    Values are Adjust Bottle success rates (\%) for the official Top~3 and
    selected open-source reference systems; higher is better. Our submission is
    highlighted. Boldface and underlining denote the best and second-best
    results among the displayed systems, respectively.}
    \label{tab:worldarena2-track2}
    \small
    \setlength{\tabcolsep}{10pt}
    \begin{tabular}{@{}L{8.2cm} r@{}}
        \toprule
        Model & Adjust Bottle \\
        \midrule
        WOVR-PLUS & \textbf{68.75} \\
        \rowcolor{BrandCyan!8}
        \textbf{DreamX-Phi-1.0-FDM-0730} & \underline{67.19} \\
        Lute & \underline{67.19} \\
        CtrlWorld & 62.50 \\
        IRASim & 61.33 \\
        RoboScape & 60.74 \\
        OpenSora & 60.16 \\
        Cosmos-Predict-2.5 (action) & 59.38 \\
        iVideoGPT & 56.25 \\
        \bottomrule
    \end{tabular}
\end{table}

On the complete 31-entry Track~1 leaderboard, our entry ranks first with an
EWMScore-P of 60.65. On Track~2, DreamX-Phi-1.0-FDM-0730
achieves a 67.19\% Adjust Bottle success rate and ties for second place in the
full snapshot. Alpha-World and FlowWAM-FiveAges complete the Track~1 Top~3,
while WOVR-PLUS leads Track~2 and Lute shares second place with our submission.
These rankings are snapshot-specific and do not represent the final challenge
standings.

\begin{figure}[H]
    \centering

    \textbf{(a) Clean scenes}
    \par\vspace{2pt}
    \includegraphics[width=\linewidth]{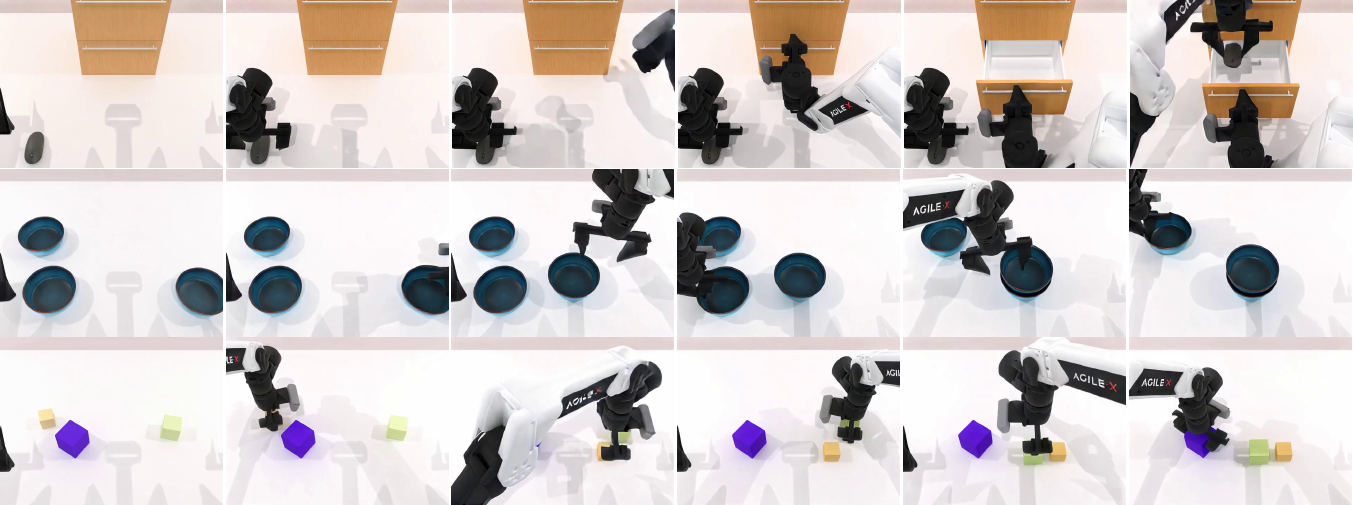}

    \vspace{6pt}
    \textbf{(b) Domain-randomized scenes}
    \par\vspace{2pt}
    \includegraphics[width=\linewidth]{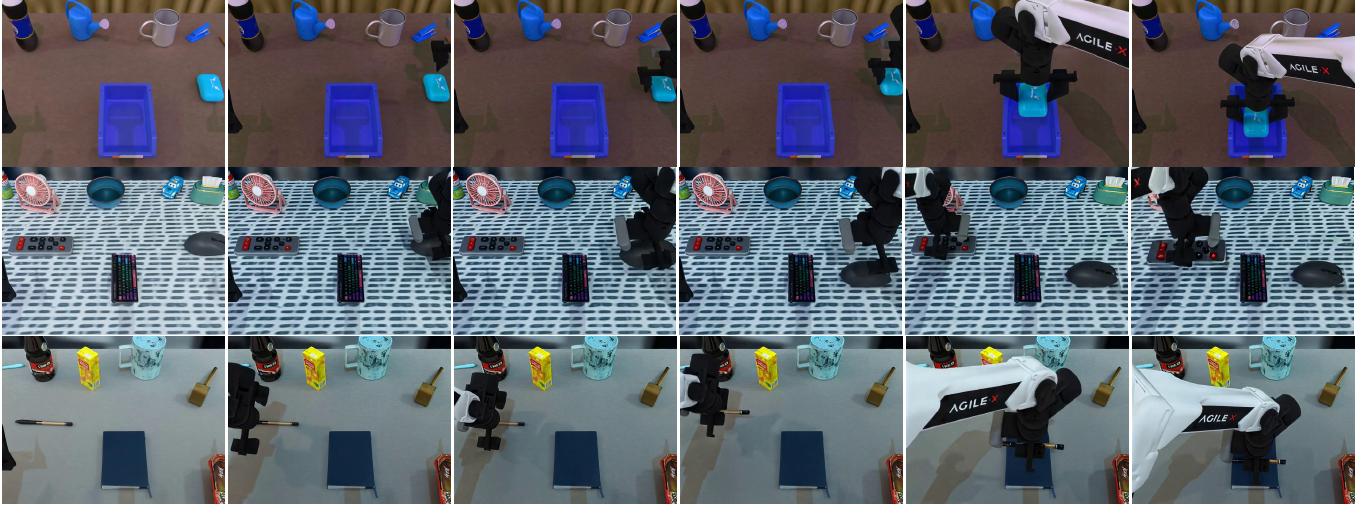}

    \caption{Qualitative WorldArena~2.0 Track~1 rollouts from
    DreamX-Phi 1.0. Each row is a predicted episode, with frames sampled in
    temporal order from left to right. (a) Standard RoboTwin~2.0 scenes.
    (b) Domain-randomized scenes with varied backgrounds, textures, lighting,
    and distractor objects.}
    \label{fig:worldarena2-track1-showcase}
\end{figure}

Figure~\ref{fig:worldarena2-track1-showcase} shows qualitative Track~1 rollouts
predicted by DreamX-Phi-1.0-FDM-0730 under both RoboTwin~2.0 evaluation settings.
The model keeps the arms, grippers, and manipulated objects coherent across the
rollout, and the same behavior holds when backgrounds, textures, lighting, and
distractor layouts are randomized.

\subsection{WorldArena~1.0 Results}

Table~\ref{tab:worldarena1-track1} places our offline WorldArena~1.0
evaluation alongside the official leaderboard Top~3 and selected open-source
reference models from the snapshot at commit \texttt{483dfcc}, dated July~15,
2026.\footnote{\href{https://huggingface.co/spaces/WorldArena/WorldArena/commit/483dfcc9a2d57974809c15d2a235c3e204f2b0f2}{WorldArena~1.0 leaderboard, commit \texttt{483dfcc}.} Baseline values are obtained from the per-model JSON files using the loader at the same commit.}
DreamX-Phi-1.0-FDM-0730 was evaluated offline on the WorldArena~1.0 Track~1 test
set and is not an entry in this pinned leaderboard snapshot.\footnote{Our
offline record reports an aggregate score of 76.88. Averaging the 15 component
values visible at four decimal places yields 76.89 after rounding; we reproduce
the reported aggregate rather than substitute the recomputed value.}

\begin{table}[H]
    \centering
    \caption{WorldArena~1.0 Track~1 comparison. Our offline evaluation of
    DreamX-Phi-1.0-FDM-0730 is shown alongside the official leaderboard Top~3 and
    selected open-source reference systems from the July~15, 2026 snapshot.
    Scores use a 0--100 scale, and EWMScore-P averages the 15 component scores.
    Higher is better; boldface and underlining denote the best and second-best
    results among the displayed systems, respectively.}
    \label{tab:worldarena1-track1}
    \scriptsize
    \setlength{\tabcolsep}{2.1pt}
    \renewcommand{\arraystretch}{0.96}

    \textbf{(a) Visual quality, motion quality, and content consistency}
    \par\vspace{2pt}
    \begin{tabular}{@{}L{2.65cm}*{9}{r}@{}}
        \toprule
        \multirow{2}{*}{Model}
        & \multicolumn{3}{c}{Visual Quality}
        & \multicolumn{3}{c}{Motion Quality}
        & \multicolumn{3}{c}{Content Consistency} \\
        \cmidrule(lr){2-4}\cmidrule(lr){5-7}\cmidrule(lr){8-10}
        & \shortstack{Image\\Quality} & \shortstack{Aesthetic\\Quality}
        & \shortstack{JEPA\\Similarity} & \shortstack{Dynamic\\Degree}
        & \shortstack{Flow\\Score} & \shortstack{Motion\\Smoothness}
        & \shortstack{Subject\\Consistency} & \shortstack{Background\\Consistency}
        & \shortstack{Photometric\\Consistency} \\
        \midrule
        \rowcolor{BrandCyan!8}
        \shortstack[l]{\textbf{DreamX-Phi-}\\
        \textbf{1.0-FDM-0730}}
        & \underline{55.72} & 40.87 & 92.73 & \textbf{88.71} & \textbf{100.00}
        & 92.22 & 82.40 & 88.96 & 10.72 \\
        UNIS & 53.94 & 40.79 & 90.60 & 73.70 & 86.02
        & \textbf{95.51} & 79.05 & 86.44 & 2.13 \\
        SisyphusWorld & 45.57 & 38.38 & \underline{96.58} & \underline{76.27} & \underline{99.82}
        & 91.42 & 82.11 & 86.52 & 5.69 \\
        BWM-Fast & 51.22 & 40.15 & \textbf{97.87} & 69.58 & 75.11
        & \underline{94.33} & 81.42 & 90.17 & 3.08 \\
        GigaWorld-0 & 48.81 & 40.15 & 38.65 & 39.45 & 30.89 & 77.34 & 73.62 & 88.15 & 12.84 \\
        Genie Envisioner & 28.18 & 26.39 & 33.81 & 32.41 & 19.91 & 69.17 & 74.53 & 87.54 & 16.04 \\
        TesserAct & 35.45 & \underline{41.30} & 39.94 & 56.49 & 26.28 & 81.80 & \textbf{84.57} & \textbf{92.82} & 20.31 \\
        RoboMaster & 36.05 & 37.87 & 29.83 & 41.74 & 16.07 & 70.91 & 82.91 & \underline{90.80} & 28.28 \\
        Vidar & 41.75 & 39.87 & 61.10 & 21.05 & 17.03 & 79.11 & 70.37 & 78.24 & 20.57 \\
        Cosmos-Predict 2.5 (text) & \textbf{65.49} & \textbf{44.33} & 46.69 & 21.62 & 17.55 & 70.93 & 66.19 & 73.98 & 35.39 \\
        Cosmos-Predict 2.5 (action) & 49.48 & 34.14 & 90.65 & 18.96 & 9.95 & 65.66 & 67.24 & 72.72 & \underline{45.21} \\
        CtrlWorld & 42.44 & 37.05 & 92.77 & 41.82 & 33.57 & 77.34 & \underline{83.56} & 90.30 & 12.88 \\
        CogVideoX & 36.23 & 36.30 & 94.84 & 31.47 & 21.77 & 73.30 & 80.97 & 88.38 & 33.79 \\
        IRASim & 41.85 & 32.86 & 93.72 & 21.60 & 11.66 & 64.31 & 72.82 & 78.69 & \textbf{45.70} \\
        \bottomrule
    \end{tabular}

    \vspace{0.45em}
    \textbf{(b) Physics adherence, 3D accuracy, controllability, and overall score}
    \par\vspace{2pt}
    \begin{tabular}{@{}L{2.65cm}*{7}{r}@{}}
        \toprule
        \multirow{2}{*}{Model}
        & \multicolumn{2}{c}{Physics Adherence}
        & \multicolumn{2}{c}{3D Accuracy}
        & \multicolumn{2}{c}{Controllability}
        & \multirow{2}{*}{\shortstack{EWM\\Score-P}} \\
        \cmidrule(lr){2-3}\cmidrule(lr){4-5}\cmidrule(lr){6-7}
        & \shortstack{Interaction\\Quality} & \shortstack{Trajectory\\Accuracy}
        & \shortstack{Depth\\Accuracy} & Perspectivity
        & \shortstack{Instruction\\Following} & \shortstack{Semantic\\Alignment}
        & \\
        \midrule
        \rowcolor{BrandCyan!8}
        \shortstack[l]{\textbf{DreamX-Phi-}\\
        \textbf{1.0-FDM-0730}}
        & 77.90 & \textbf{58.98} & 93.17 & 96.30
        & 84.92 & 89.68 & \textbf{76.88} \\
        UNIS & \textbf{87.30} & 41.89 & 85.25 & \textbf{98.84}
        & \textbf{93.86} & 89.35 & \underline{73.64} \\
        SisyphusWorld & 71.98 & 44.58 & \textbf{94.85} & 90.34
        & 82.68 & 89.18 & 73.06 \\
        BWM-Fast & \underline{79.88} & 44.89 & 86.41 & 97.14
        & \underline{90.22} & 89.15 & 72.71 \\
        GigaWorld-0 & 56.20 & 17.07 & 63.78 & 78.06 & 56.14 & 86.64 & 53.85 \\
        Genie Envisioner & 22.74 & 2.63 & 86.83 & 54.32 & 20.36 & 85.98 & 44.06 \\
        TesserAct & 62.70 & 16.39 & 73.85 & 84.26 & 66.80 & 88.68 & 58.11 \\
        RoboMaster & 54.80 & 12.35 & 83.77 & 77.94 & 51.14 & 87.78 & 53.48 \\
        Vidar & 60.84 & 21.26 & 79.77 & 81.88 & 61.54 & 88.46 & 54.86 \\
        Cosmos-Predict 2.5 (text) & 65.14 & 11.60 & 70.71 & \underline{98.22} & 64.70 & 86.34 & 55.93 \\
        Cosmos-Predict 2.5 (action) & 60.46 & 27.49 & 90.35 & 84.68 & 60.54 & \textbf{89.86} & 57.83 \\
        CtrlWorld & 62.62 & \underline{48.20} & 93.25 & 83.66 & 67.68 & 88.68 & 63.72 \\
        CogVideoX & 66.96 & 34.79 & 91.09 & 85.46 & 75.58 & \underline{89.70} & 62.71 \\
        IRASim & 62.76 & 35.92 & \underline{93.50} & 84.16 & 65.40 & 89.43 & 59.63 \\
        \bottomrule
    \end{tabular}
\end{table}

The fixed WorldArena~1.0 leaderboard is led by UNIS (73.64), followed by
SisyphusWorld (73.06) and BWM-Fast (72.71). In the same 15-metric format, our
offline DreamX-Phi-1.0-FDM-0730 result reaches 76.88, 3.24 points above the leading
official entry in this snapshot.

\section{Limitations}
\label{sec:limitations}

Our evaluation is limited to WorldArena and RoboTwin, with Track~2 covering
only the Adjust Bottle task, so generalization to other tasks, embodiments, and
real robots remains unverified. The leaderboard scores evaluate the full system
and therefore do not isolate the contribution of individual components.
Finally, \model{} predicts videos from externally provided actions rather than
generating actions itself. Track~2 shows that the model can serve as a rollout
environment for training a separate policy, but it does not evaluate \model{}
as a closed-loop controller.

\section{Conclusion}
\label{sec:conclusion}

We presented \modelversion{}, an action-conditioned video world model for
robotic manipulation. The model addresses a central challenge in bimanual video
prediction: a generated rollout must follow the prescribed motion of each arm
while preserving scene geometry and the state of the manipulated object. To
this end, \model{} combines arm-specific $\mathrm{SE}(3)$ conditioning with
auxiliary depth and object-centric supervision. On the August~12, 2026
WorldArena~2.0 snapshot, our submission ranked first among 31 Track~1 entries
with an EWMScore-P of 60.65. In Track~2, a policy trained using the submitted
world model as its rollout environment achieved a 67.19\% success rate on
Adjust Bottle, tying for the second-highest score in the same snapshot. These
results demonstrate strong system-level performance in video prediction and
world-model-based policy training, while matched ablations are still needed to
quantify the contribution of each component.

\section{Future Work}
\label{sec:future_work}

The current \modelversion{} is formulated as a Forward Dynamics Model
(FDM), which predicts future observations from an externally
provided action sequence and does not generate actions itself. We will
extend this framework to other model formulations. In particular, we plan to
develop a joint World Action Model that generates future video and robot
action trajectories together. Training these outputs jointly should help align
each proposed action with its predicted visual consequences. We will evaluate
the resulting model in terms of video quality, action accuracy, action--video
consistency, and closed-loop task success.

\section*{Authors}

\paragraph{Team Members.}
\emph{Team members are listed alphabetically by last name (and by first name
where last names are identical). The ordering does not indicate relative
contributions.}

Rui Chen, Xiangxiang Chu, Geng Li, Jifan Li, Qingfeng Shi, Datao Tang, Jing Tang, Jun Wang, Pengfei Zhang.

\bibliographystyle{plainnat}
\bibliography{references}

\end{document}